\documentclass[10pt,twocolumn,letterpaper]{article}

\usepackage{cvpr}
\usepackage{times}
\usepackage{epsfig}
\usepackage{graphicx}
\usepackage{amsmath}
\usepackage{amssymb}

\usepackage[breaklinks=true,bookmarks=false]{hyperref}

\usepackage{booktabs}
\usepackage{float}

\usepackage{multirow}
\usepackage{subfigure}
\usepackage{bm}

\usepackage[T1]{fontenc}

\usepackage{rotating}
\usepackage{array}

\cvprfinalcopy 

\def\cvprPaperID{****} 

\begin{document}

\title{A Large-scale Attribute Dataset for Zero-shot Learning }
\author{Bo Zhao$^1$, Yanwei Fu$^2$, Rui Liang$^3$, Jiahong Wu$^3$, Yonggang Wang$^3$, Yizhou Wang$^1$ \\
$^1$ National Engineering Laboratory for Video Technology, \\ Key Laboratory of Machine Perception (MoE),\\ Cooperative Medianet Innovation Center, Shanghai, \\ School of EECS, Peking University, Beijing, 100871, China\\
$^2$ School of Data Science, Fudan University\\
$^3$ Sinovation Ventures\\
bozhao, yizhou.wang@pku.edu.cn, yanweifu@fudan.edu.cn\\
liangrui, wujiahong, wangyonggang@chuangxin.com
}
\maketitle


\begin{abstract}

Zero-Shot Learning (ZSL) has attracted huge research attention over the past few years; it aims to learn the new concepts that have never been seen before. In classical ZSL algorithms, attributes are introduced as the intermediate semantic representation to realize the knowledge transfer from seen classes to unseen classes. Previous ZSL algorithms are tested on several benchmark datasets annotated with attributes. However, these datasets are defective in terms of the image distribution and attribute diversity. In addition, we argue that the ``co-occurrence bias problem'' of existing datasets, which is caused by the biased co-occurrence of objects, significantly hinders models from correctly learning the concept. To overcome these problems, we propose a Large-scale Attribute Dataset (LAD). Our dataset has 78,017 images of 5 super-classes, 230 classes. The image number of LAD is larger than the sum of the four most popular attribute datasets. 359 attributes of visual, semantic and subjective properties are defined and annotated in instance-level. We analyze our dataset by conducting both supervised learning and zero-shot learning tasks. Seven state-of-the-art ZSL algorithms are tested on this new dataset. The experimental results reveal the challenge of implementing zero-shot learning on our dataset.

\end{abstract}

\section{Introduction}

Humans can distinguish more than 30,000 basic level concepts and many
more subordinate ones \cite{biederman1987recognition}, while existing
deep neural networks \cite{simonyan2014very,szegedy2015going,he2016deep}
can only classify thousands of objects. It is expensive to collect
the labelled data sufficiently to train deep neural networks for all
classes. Human beings, in contrast, can leverage the semantic knowledge
(e.g., textual descriptions) to learn the novel concepts that
ones have never seen before. Such the ``learning to learn'' ability
inspires the recent study of zero-shot learning (ZSL) \cite{palatucci2009zero},
which targets at identifying novel classes without any training examples.
In practice, the ZSL is achieved via inferring the intermediate semantic
representations that may be shared both by the seen and unseen concepts.
In particular, the middle-level semantic representations (e.g. attributes)
are utilized to make connections between the low-level visual features
and high-level class concepts.

Many different semantic representations have been investigated, such
as semantic attributes \cite{lampert2014attribute}, word vectors
\cite{mikolov2013distributed} and gaze embeddings \cite{karessli2017gaze}.
Though they have to be manually labeled, semantic attributes have
been most widely used due to the good merits of ``name-ability\textquotedblright{}
and ``discriminativeness\textquotedblright . Additionally, the attributes
can also facilitate the zero-shot generation (ZSG) \cite{yan2016attribute2image,reed2016generative,yin2017semi,mathieu2016disentangling},
which aims to \textcolor{black}{{} generate the images of unseen classes with novel semantic representations.}

The image datasets annotated with attributes such as Caltech-UCSD
Birds-200-2011 (CUB) \cite{WahCUB_200_2011}, SUN Attributes (SUN)
\cite{xiao2010sun}, aPascal\&aYahoo (aP\&aY) \cite{farhadi2009describing}
and Animals with Attributes (AwA) \cite{farhadi2009describing}, are
widely used as the testbed for ZSL algorithms. However, the total
number of images and attributes of these dataset are too limited to
train from the scratch the state-of-the-art deep models \emph{namely},
VGGs \cite{VGG}, ResNets \cite{ResidualNet} and DenseNets \cite{denseNet}.

\begin{figure}
\centering{}\includegraphics[width=0.46\textwidth]{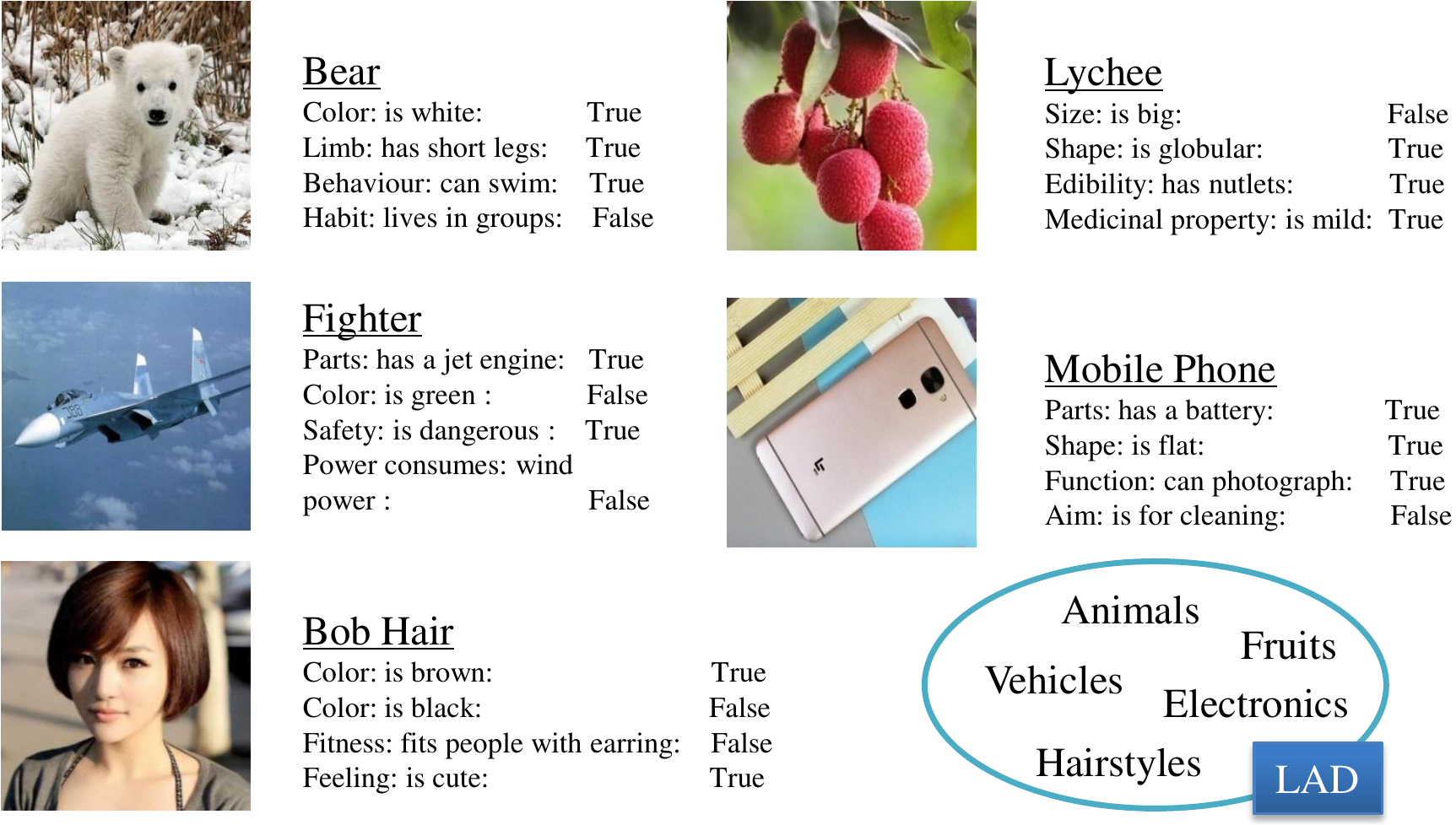} \caption{\label{fig:example_dataset}
\textcolor{black}{The overview of the proposed LAD dataset. It includes
230 classes belonging to five super-classes (domains). Labels, bounding boxes and attributions are annotated. The upper two attributes are visual attributes, while the bottom two are semantic attributes.}}
\end{figure}

Furthermore, there exist several additional issues with these attribute
datasets. (1) The categories and images of these datasets may be highly
related to ImageNet dataset (used in ILSVRC 2010/2012).
\textcolor{black}{In ZSL scenario,
it is thus less desirable to directly utilize the deep models pre-trained on ILSVRC 2010/2012 as the feature extractors,
 which may include the images of novel unseen classes from these datasets \cite{zsl_ugly}.
} (2) These datasets (CUB and SUN) may focus on each specific
visual domain; and yet the datasets for the common object (aP/aY)
and animal (AwA) domains do not really have sufficient fine-grained
classes to validate the knowledge transfer in zero-shot scenario.

Additionally, there exists serious ``co-occurrence bias'' in these
datasets. For example, the ``person'' objects occur in many AwA
classes with a high frequency. Such correlation may be implicitly
learned and utilized as the cues of identifying zero-shot classes.
Specifically, suppose we want to identify two unseen classes \textendash{}
lion and dog. Essentially, these two classes share many common attributes,
such as ``four legs'', ``has fur'' and so on. Actually, even visually
some kind of dog (\emph{e.g.}, Tibetan mastiff) is very similar to
lion. However, the zero-shot algorithms may easily identify the dog
class by only detecting whether ``person'' objects are present in
the image since high protion of ``person'' and "dog" objects are co-occurrence  in the dog class of this dataset.
 Such ``co-occurrence'' is caused by the way of how we
construct the dataset; and thus can be taken as one type of bias.
The algorithms implicitly utilize this correlation may be limited
to generalize to other domains which do not have such type of correlation.

To alleviate these problems of existing datasets, we are making efforts
of contributing the new attribute dataset \textemdash{} Large-scale
Attribute Dataset (LAD) to the community. We design a novel label
list and collect images from different sources, in order to get more
new classes and images different from existing datasets. Except for
low-level visual attributes (\emph{e.g}. colors, sizes, shapes), we
also provide many attributes about semantic and subject visual properties
\cite{fu2016robust}. For example, as illustrated in Fig. \ref{fig:example_dataset},
we annotate attributes of diets and habits for ``animals''; edibility
and medicinal property for ``fruits''; safety and usage scenarios
for ``vehicles''; functions and usage mode for ``electronics'';
human feelings for ``hairstyles''. We cluster classes into several
super-classes. Each super-class can be viewed as a fine-grained subset,
and the attributes are designed for each super-class. Then, the knowledge
transfer between fine-grained classes are feasible.

To break the co-occurrence among objects, we collect the images with
only single (foreground) object.  Overall, we constructed a new attribute
dataset which contains 78,017 images from 230 classes. These classes
are from 5 different visual domains (super-class), including animals,
fruits, vehicles, electronics, and hairstyles. 359 visual, semantic
and subjective attributes are annotated for randomly selected 20 images
per class.

The central contribution of our paper is to propose a large-scale
attribute dataset which is larger than the sum of the four most popular
datasets in ZLS. In addition, several baseline experiments on supervised
learning and zero-shot learning are conducted on this new dataset.
Seven state-of-the-art methods are re-implemented, and the zero-shot
recognition accuracies are reported as the baselines. The LAD aims
at being a reasonable testbed for evaluating the zero-shot learning
algorithms. The LAD \footnote{Dataset Download: \href{https://github.com/PatrickZH/A-Large-scale-Attribute-Dataset-for-Zero-shot-Learning}
{https://github.com/PatrickZH/A-Large-scale-Attribute-Dataset-for-Zero-shot-Learning}}
has been used as the benchmark dataset for
Zero-shot Learning Competition in AI Challenger\footnote{Competition Website: \href{https://challenger.ai/}{https://challenger.ai/}}.

The rest of the paper is organized in the following way. Sec. \ref{sec:Related-Work}
discusses some previous works related to zero-shot learning and attribute
datasets. We elaborate how to construct the LAD in Sec. \ref{sec:Dataset-Construction}.
We present a set of the data splits for evaluating different ZSL algorithms
on the LAD in Sec. \ref{sec:Data-Split}. The experiments and the
corresponding discussion are detailed in Sec. \ref{sec:Experiments}.
Sec. \ref{sec:Conclusion} concluded the whole paper.

\section{Related Work\label{sec:Related-Work}}

\subsection{Attributes}

Attributes are used to describe the property of one instance or a
class \cite{farhadi2009describing,lampert13AwAPAMI}. The attribute
can serve as an intermediate representation for knowledge transfer
between seen and unseen classes; and thus facilitates a wide range
of tasks in transfer learning scenarios, e.g., zero-shot recognition
and zero-shot generation. In term of the ``name-ability'' and ``discriminativeness'',
the attributes can be roughly categorized into several groups including,

\vspace{0.1in}

\noindent \textbf{User-defined Attributes}. These attributes are defined
by human experts \cite{lampert13AwAPAMI} or knowledge ontology \cite{rohrbach2010semantic_transfer}.
According to different domains and tasks, the distinctive attributes
can be manually defined, such as, person attributes \cite{wang2011clothesattrib,moon_attrb,wang2016walk,datta2011face_attrib},
biological traits (\emph{e.g.}, age and gender) \cite{facial_attrb_icmr},
product attributes (\emph{e.g.}, size, color, price) and 3D shape
attributes \cite{3D_shape_attribute} and so on. These attributes
have to reflect the intrinsic properties of one instance/class (i.e.
``name-ability''), but  is also useful in knowledge transfer for
recognition/generation tasks (i.e. ``discriminativeness'').

\vspace{0.1in}

\noindent \textbf{Data-driven Attributes}. Due to the expense of exhaustive
annotation of user-defined attributes, some methods \cite{liu2011action_attrib,yanweiPAMIlatentattrib,video_story_1shot}
mine the data-driven attributes. These attributes usually have good
``discriminativeness'' and yet usually has less ``name-ability'',
since it is generally difficult to give the names of attributes discovered
from the data.

\vspace{0.1in}

\noindent \textbf{Generalized Attributes}. The attributes can be generalized
in other types of semantic representations, such as semantic word
vector of each class name \cite{DeviseNIPS13,RichardNIPS13,norouzi2013zero}
and gaze embeddings \cite{karessli2017gaze}. In general, these generalized
attributes are trained and extracted from the other domains. For example,
word2vec models \cite{mikolov2013distributed} are trained on huge
texts. Thus the generalized attributes may not fit visual representations
well, but they are complementary to user-defined attributes \cite{transductiveEmbeddingJournal}.

Currently, the user-defined visual attributes have been well investigated.
For instance, 312 colors, shapes, patterns, sizes and lengths of different
bird parts are annotated in CUB dataset \cite{WahCUB_200_2011}.
\textcolor{black}{To further push the deeper research of attribute learning, it is essential
to generalize visual attributes to more diverse and versatile semantic
representations. To this end, the LAD introduces human-concerned semantic
attributes (e.g. the habits of animals) and the subjective visual properties (e.g. the feelings about hairstyles) \cite{parikh2011relativeattrib,fu2016robust}.
}

\subsection{Zero-shot Learning}

In this paper we focus on two zero-shot learning tasks, \emph{namely},
zero-shot recognition and zero-shot generation. Please refer to \cite{yanwei2017spm}
for a more detailed review.

\vspace{0.1in}

\noindent \textbf{Zero-shot Recognition (ZSR). } ZSR has attracted
significant research attention in the past few years. Extensive efforts
and previous works can be roughly divided into three groups. (1) Direct
embedding from visual space to semantic space (or reverse embedding) \cite{palatucci2009zero,lampert2014attribute}.
It learns a mapping function from the visual feature space to the
semantic embedding space by auxiliary training data; the learned mapping
function is directly applied to project the unseen testing images
into semantic space and match against the prototype of the novel class/concept.
(2) Learning the joint embedding space \cite{akata2013label,xian2016latent,romera2015embarrassingly}.
Both the image features and semantic embeddings are jointly projected
into a new embedding space. For each given testing unseen image, its
label is predicted according to the distance to unseen semantic embeddings
in the new space. (3) Transferring structural knowledge from semantic
space to visual space \cite{naha2015zero,changpinyo2016synthesized,zhao2017zero}.
The structural knowledge is learned in semantic space, and then transferred
to the visual space for synthesizing the visual instances or classifiers
of unseen classes.

\vspace{0.1in}

\noindent \textbf{Zero-shot Generation (ZSG).} In recent years, zero-shot
generation methods synthesize images conditioned on attributes/texts
using generative models. \cite{yan2016attribute2image} presented
a successful trial to synthesize natural images of birds and faces.
They choose Conditional Variational Auto-Encoder as the basic model
and then disentangle the foreground and background by introducing
a layer representation. PixelCNN is utilized to model images conditioned
on labels, tags or latent embeddings \cite{van2016conditional}. However,
new images can be synthesized only based on existing labels, latent
embeddings or the linear interpolations of them. Some methods \cite{reed2016generative,yin2017semi,zhang2016stackgan,perarnau2016invertible}
based on conditional GAN \cite{mirza2014conditional} have been proposed
to generate images with unseen attribute/text representation. In \cite{reed2016generative},
the encoding of text is used as the condition in both the generator
and discriminator by concatenating the random noise and image feature
maps. \cite{yin2017semi} proposed a novel model to synthesize and
edit facial images. The semi-latent facial attribute space, which
includes both learned latent attributes and user-defined attributes,
are leveraged as the conditional input of GAN. Note that most of these methods
focus on the image generation with particular visual attributes or
descriptions, \emph{e.g.} colors, parts of faces, flowers and birds.
However, in this paper, we try a more difficult task, \emph{i.e.,}
manipulating semantic attributes and generating images with abstract
attributes as discussed in Sec. \ref{subsec:Zero-shot-Generation}.

\subsection{Image-based Attribute Datasets }

Several datasets are repurposed by annotating attributes in order
to evaluate the zero-shot learning algorithms. These datasets include
Caltech-UCSD Birds-200-2011 (CUB) \cite{WahCUB_200_2011}, SUN Attributes
(SUN) \cite{xiao2010sun}, aPascal\&aYahoo (aP\&aY) \cite{farhadi2009describing},
Animals with Attributes (AwA) \cite{farhadi2009describing}, Public
Figures Face Database (PubFig) \cite{kumar2009attribute}, Human Attributes
(HAT) \cite{sharma2011learning} and Unstructured Social Activity
Attribute (USAA) \cite{fu2012attribute}. Essentially, any dataset,
if labeled with attributes or word vectors, can be used to evaluate
the ZSR/ZSG algorithms. The statistics of the most popular four attribute
datasets are shown in Tab. \ref{tab:statistics}.

\textcolor{black}{ As aforementioned, existing benchmarks have three main drawbacks.
(1) The categories and images of existing attribute datasets may be
highly reused in ILSVRC 2010/2012 which are frequently used to pre-train the deep feature extractors. Hence,
the image feature extractors may have seen many testing (``unseen'') classes. (2) The categories are not fine-grained
enough. For example, aP/aY contains only 32 coarse-grained
categories. As aforementioned, those datasets do not have sufficient fine-grained classes to validate the knowledge transfer in ZSL. (3) There exists serious co-occurrence bias in these datasets.
AwA and aP/aY contain images with multiple foreground objects; however
every image only has a single label. Some objects have a biased co-occurrence with others in particular classes. For instance, 30\% classes in
AwA have more than 10\% images containing ``person''. Even, in ``ox''
and ``horse'' classes, the co-occurrence frequency is greater than
35\%. To overcome these three drawbacks, we introduce a new benchmark as the testbed of
zero-shot learning.}

\begin{table*}
\centering{}%
\begin{tabular}{c|c|c|c|c|c}
\hline
 & {\small{}{}{}{}{}{}{}LAD}  & {\small{}{}{}{}{}{}{}CUB-bird}  & {\small{}{}{}{}{}{}{}SUN}  & {\small{}{}{}{}{}{}{}aP/aY}  & {\small{}{}{}{}{}{}{}AwA} \tabularnewline
\hline
{\small{}{}{}{}{}{}{}Images}  & {\small{}{}{}{}{}{}{}}\textbf{\small{}{}78,017}{\small{} } & {\small{}{}{}{}{}{}{}11,788}  & {\small{}{}{}{}{}{}{}14,340}  & {\small{}{}{}{}{}{}{}15,339}  & {\small{}{}{}{}{}{}{}30,475} \tabularnewline
\hline
{\small{}{}{}{}{}{}{}Classes}  & {\small{}{}{}{}{}{}{}230}  & {\small{}{}{}{}{}{}{}200}  & {\small{}{}{}{}{}{}{}}\textbf{\small{}{}717}{\small{} } & {\small{}{}{}{}{}{}{}32}  & {\small{}{}{}{}{}{}{}50} \tabularnewline
\hline
{\small{}{}{}{}{}{}{}Bounding Box}  & {\small{}{}{}{}{}{}{}Yes}  & {\small{}{}{}{}{}{}{}Yes}  & No  & {\small{}{}{}{}{}{}{}Yes}  & {\small{}{}{}{}{}{}{}No} \tabularnewline
\hline
{\small{}{}{}{}{}{}{}Attributes}  & {\small{}{}{}{}{}{}{}}\textbf{\small{}{}359}{\small{} } & {\small{}{}{}{}{}{}{}312}  & {\small{}{}{}{}{}{}{}102}  & {\small{}{}{}{}{}{}{}64}  & {\small{}{}{}{}{}{}{}85} \tabularnewline
\hline
{\small{}{}{}{}{}{}{}Annotation Level}  & {\small{}{}{}{}{}{}{}20 ins./class}  & {\small{}{}{}{}{}{}{}instance}  & {\small{}{}{}{}{}{}{}instance}  & {\small{}{}{}{}{}{}{}instance}  & {\small{}{}{}{}{}{}{}class} \tabularnewline
\hline
\end{tabular}{\small{}{}{}{}{}{}\caption{{ {}{}{}{}{}\label{tab:statistics} Statistics and comparison
of different datasets used in zero-shot learning. }}
}
\end{table*}

\section{Dataset Construction\label{sec:Dataset-Construction}}

In this paper, we construct a Large-scale Attribute Dataset (LAD)
which can serve as the testing bed for zero-shot learning algorithms.
The construction process of LAD can be divided into four steps, \emph{namely},
the definition of classes and attributes (Sec. \ref{subsec:Label-and-Attribute}),
image crawling (Sec. \ref{subsec:Image-Crawling}), data preprocessing
(Sec. \ref{subsec:Preprocessing}) and data annotation (Sec. \ref{subsec:Annotation}).

\subsection{Definition of classes and attributes \label{subsec:Label-and-Attribute}}

\begin{figure*}
\centering{}\includegraphics[width=0.66\textwidth]{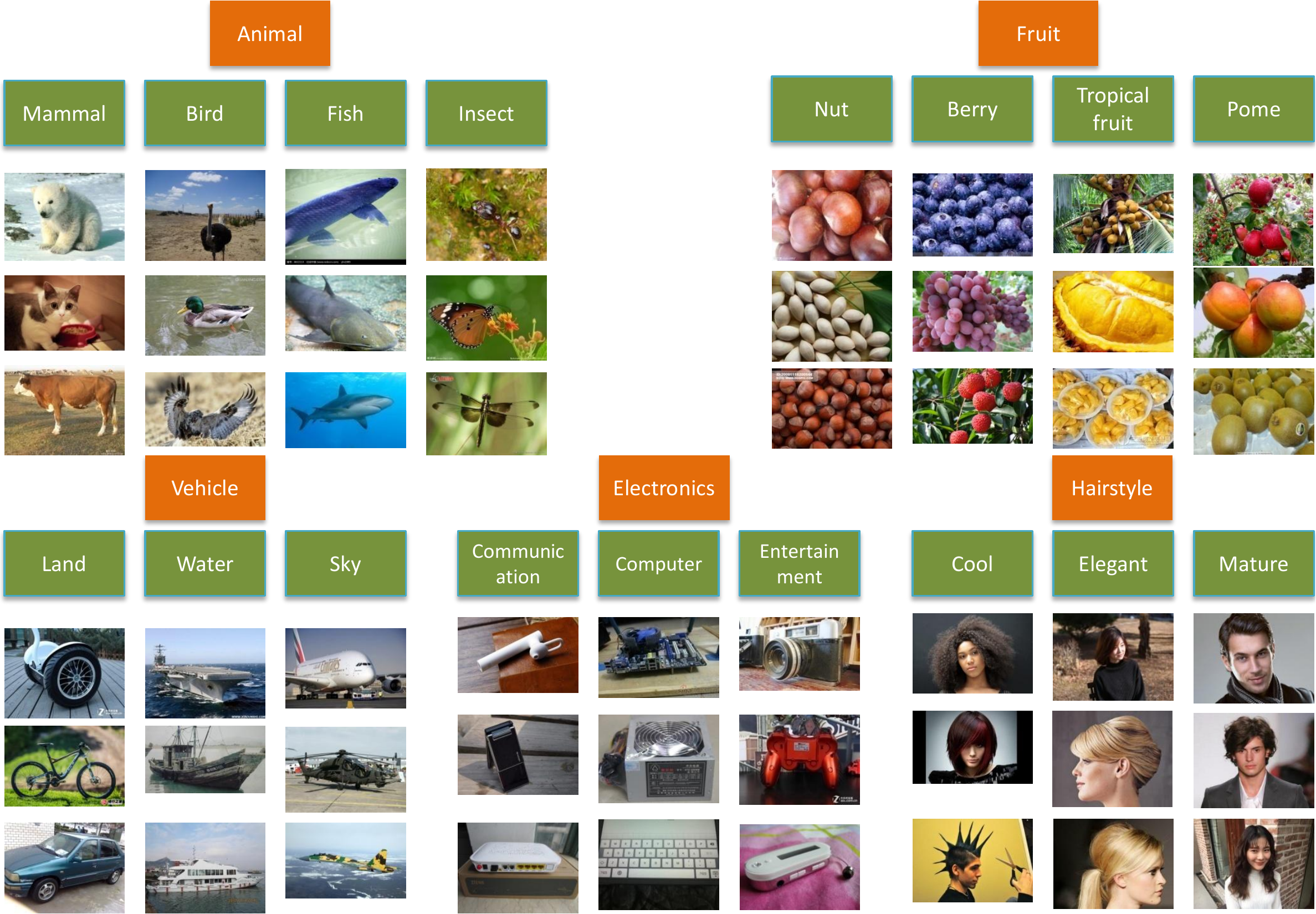} \caption{\label{fig:classtree} The hierarchical structure of
super-classes (domains) and some example classes. }
\end{figure*}

\noindent \textbf{Classes.} It is of central importance to well define
the classes and attributes of an attribute dataset. In general, we expect the LAD have
more common classes, and yet fewer shared classes with ImageNet
dataset (ILSVRC 2010/2012) \cite{deng2009imagenet}. It is nontrivial, since ImageNet
dataset is built upon the well-known concept ontology \textendash{}
WordNet \cite{WordNet_1995Miller}. Critically, we define the classes
from five domains (super-classes), \emph{namely}, animals, fruits,
vehicles, electronics and hairstyles. Animals and fruits are natural
products, while vehicles and electronics are artificial products.
We choose 50 popular classes for animals, fruits, vehicles, electronics.
The hairstyle super-classes include the 30 mostly popular Asian and
Western hairstyles. All these classes are selected as less overlapped
with the WordNet ID of ILSVRC 2010/2012 dataset. In particular, some classes
(\emph{e.g}., the ``fauxhawk'' and ``mullet'' in the hairstyle super-class)
have only recently been collected and annotated to the community \cite{hairstyle}.
Some example images of different classes are shown in the tree structure
of Fig. \ref{fig:classtree}.

\vspace{0.1in}

\noindent \textbf{Attributes.} Considering the huge diversity of LAD,
we design the attribute list for each super-class; more specifically,
we define $123$, $58$, $81$, $75$ and $22$ attributes for animals,
fruits, vehicles, electronics and hairstyles respectively. The defined
attributes include visual information ( \emph{e.g}. color, shape,
size, appearance, part, and texture), or the visual semantic information
such as ``whether an type of animal eats meat?'', or subjective visual
properties \cite{fu2016robust}, \emph{e.g.}, ``whether the hairstyle
gives the feeling of cute?''. Such a type of attribute definition
will facilitate designing zero-shot learning algorithms by transferring
various information \textendash{} visual information, semantic information
and subjective visual properties. Additionally, even more wider types
of attributes have been considered here; for example, we annotate
attributes of diets and habits for the animal super-class, edibility
and medicinal property for the fruit super-class, safety and usage
scenarios for the vehicle super-class, functions and usage mode for
the electronic super-class. The knowledge of such attributes is referring
to \emph{Wikipedia}. The full list of attributes will be provided
in the supplementary material.

\subsection{Image Crawling\label{subsec:Image-Crawling}}

We gather the images of each defined class by using the popular search
engines\emph{, e.g. Baidu} and \emph{Google}. Specifically, to efficiently
search enough images, the class names by different languages (\emph{e.g.}
English and Chinese) have been used in the search engines. We also
use synonyms and determiners to obtain better search results. By this
mean, for each class we crawled about 1,000 images with public licenses.

\subsection{Preprocessing \label{subsec:Preprocessing}}

\noindent \textbf{Initial Preprocessing.} The raw crawled images are very noisy. Huge human efforts are devoted to clean up
the crawled images. Specifically, we manually remove those images
of low quality (\emph{e.g.}, low-resolution, or large watermark).
Also for each class, those duplicated or unrelated noisy images are
also manually pruned.

\vspace{0.1in}

\noindent \textbf{Removing Co-occurrence Bias. }Considering that the
co-occurrence bias of one dataset mostly comes from the co-existent
objects with the proposed object class. For example, many images of
the animals in AwA contain the ``person'' object. To avoid such
cases, we prefer the images with iconic view of each class.
\textcolor{black}{ Particularly,
we take as the background, the sky, lakes, land, trees, buildings,
blur objects and tiny objects; and those images have more than one foreground
object of iconic view would be discarded.}

\subsection{Annotation\label{subsec:Annotation}}

\noindent \textbf{Class Annotation.} We also need to further annotate
the preprocessed images. In particular, we remove those images whose
foreground objects mismatch the class name/label. We finally obtain $78k$ images of all five super-classes as shown
in Tab. \ref{tab:statistics}. We also annotate the bounding box of
each foreground object.

\vspace{0.1in}

\noindent \textbf{Attribute Annotation.} According to the attribute list defined in Sec. \ref{subsec:Label-and-Attribute},
we annotate instance-level attributes for selected images. Specifically, we randomly select 20 images per class to annotate
attributes. The class-level attributes can be computed
as the mean values of the attributes of 20 images.

\begin{figure*}
\centering \subfigure[Statistics of image numbers.]{ \label{fig:statistics.sub.1}
\includegraphics[width=0.48\textwidth]{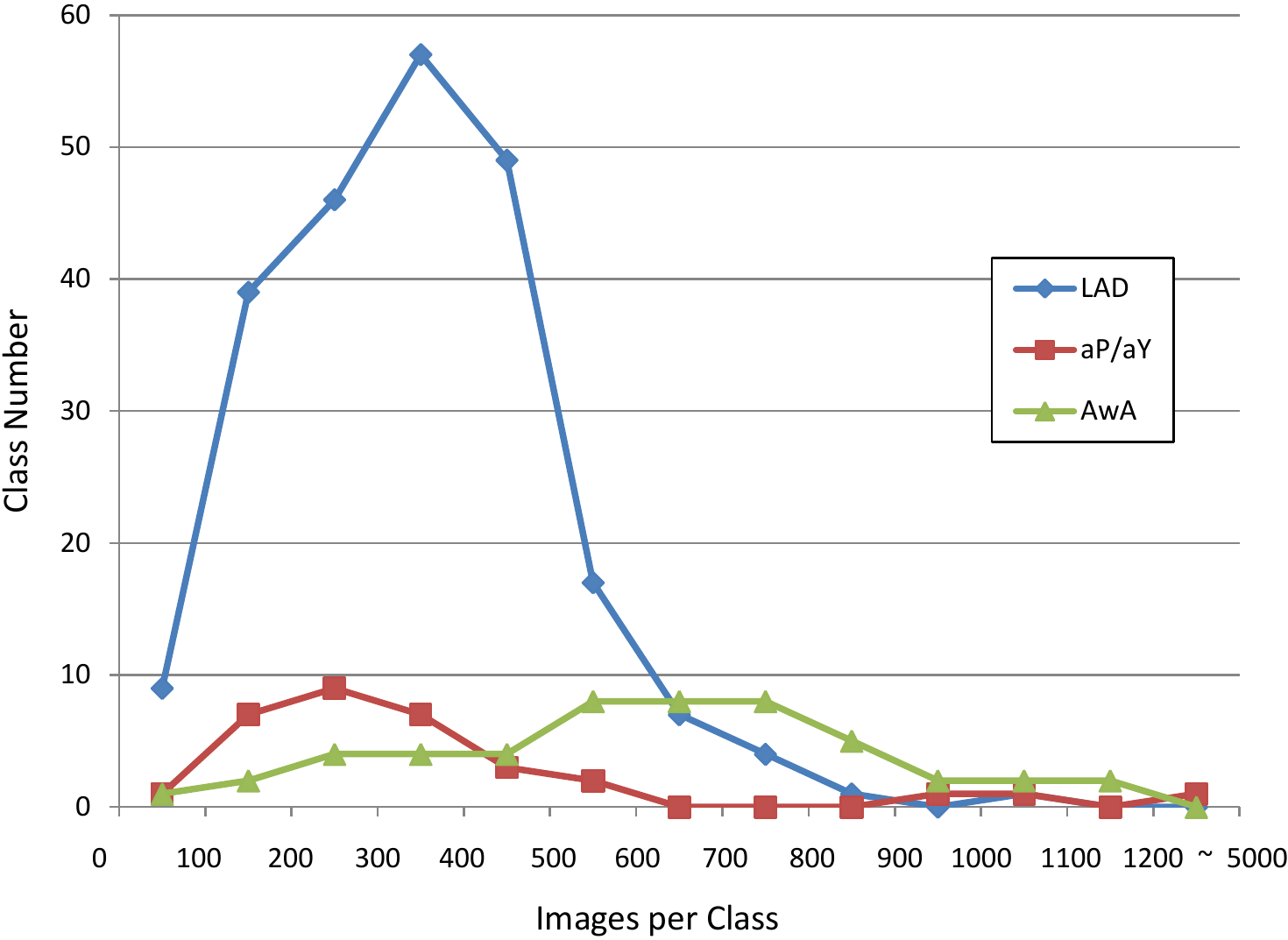}} \subfigure[Statistics of class and attribute numbers.]{
\label{fig:statistics.sub.2} \includegraphics[width=0.48\textwidth]{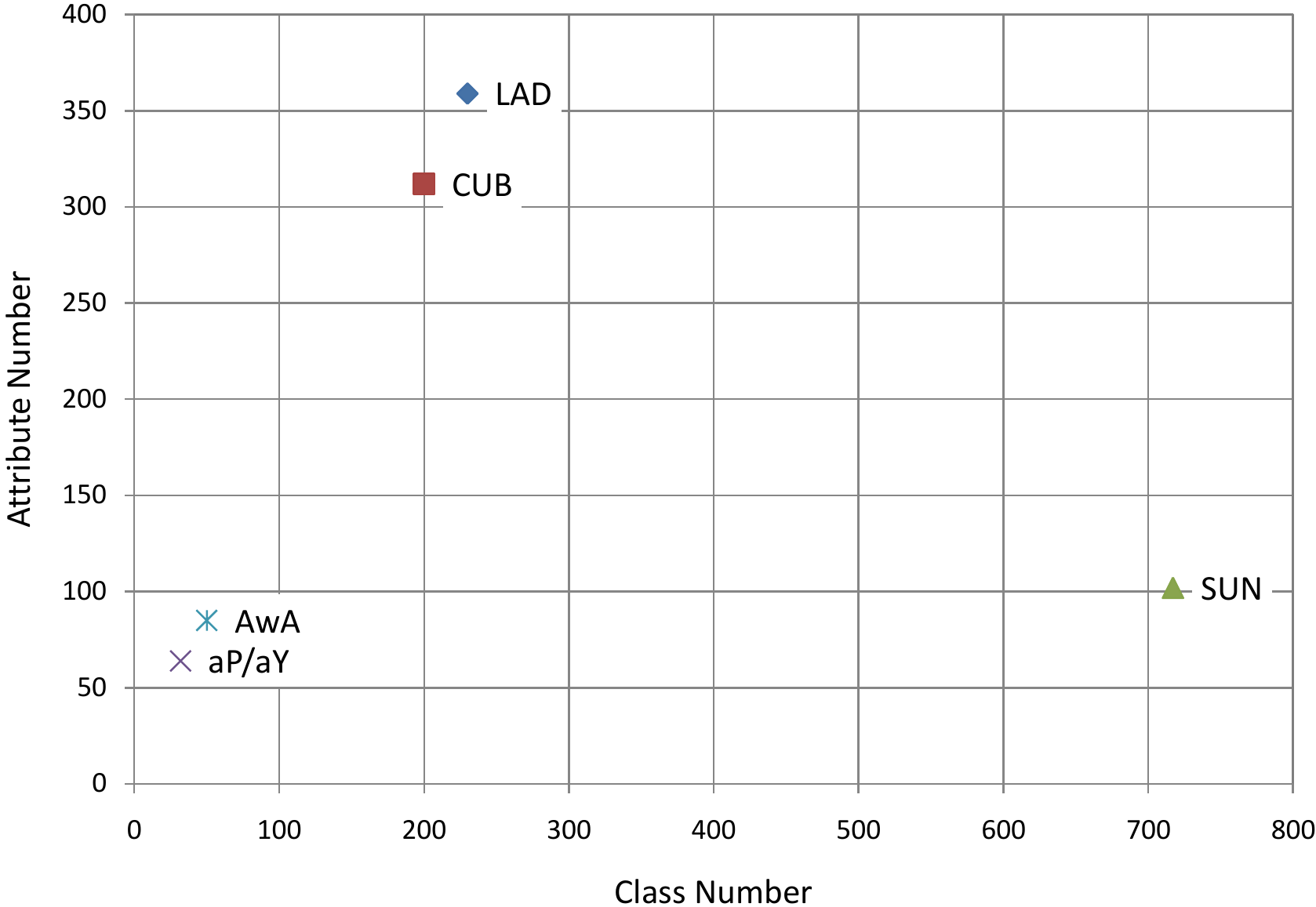}}
\caption{{}{}{}{}{Statistics of image, class and attribute numbers of
different attribute datasets.}}
\label{fig:statistics}
\end{figure*}

\subsection{Statistics}

\noindent \textbf{Total Images.} Our LAD dataset contains 78,017 images.
As shown in Fig. \ref{fig:statistics.sub.1}, we compare the distribution
of image number per class with the AwA and aP/aY datasets. It shows
that most classes of LAD have 350 images whilst the AwA and aP/aY
have around 650 and 250 images per classes respectively. In particular,
AwA has 30,475 images from 50 animals with 85 class-level attributes;
and aP/aY includes 15,339 images of 32 classes with 64 instance-level
attributes. The area under the curve indicates the total images of
each dataset. It means that our LAD is much larger than the AwA and
aP/aY datasets.

\vspace{0.1in}

\noindent \textbf{Classes and Attributes.} \textcolor{black}{Fig. \ref{fig:classtree} provide a hierarchical view of part of classes in our dataset.
We can find that every super-class is fine-grained. }
We also compare the total class and attribute numbers of  LAD, CUB, SUN,
aP/aY and AwA datasets in Fig. \ref{fig:statistics.sub.2}.
Our LAD contains the 359 attributes which is much larger than the attribute
number of other datasets. The sheer volume of annotated attributes
essentially provide a good testing bed for the zero-shot learning
algorithms. Furthermore, we also introduce the subjective attributes
of human hairstyle classes as illustrated in Fig. \ref{fig:example_dataset}.
Comparably, the SUN dataset has 717 classes and yet only 14,340 images
and 102 annotated attributes. The CUB-200 2011 bird dataset has 312
attributes, which however only focus on the visual information such
as colors, shapes, patterns, sizes and lengths of the birds. In contrast,
our LAD introduces the attributes of visual semantic information and
subjective visual properties \cite{fu2016robust}; and we argue that
these attributes are much richer semantic representation and potentially
can be better used for knowledge transfer in testing the zero-shot
algorithms.

We calculate the ratio of shared classes between different attribute
datasets and ImageNet dataset. In particular, we use the
competition data in ILSVRC 2012, because most deep feature extractors are trained on ILSVRC 2012.
For each dataset, we count the number of class names which exist in the WordNet ID
of ILSVRC 2012. The ratio is calculated by dividing the total
class number of each dataset. As shown in Tab. \ref{tab:overlap},  our dataset has
only 43.48\% overlap ratio with ILSVRC 2012 which is significantly
lower than AwA (76.00\%) and aP/aY (86.96\%).

\begin{table}
\centering{}
\begin{tabular}{c|c|c|c}
\hline
Dataset  & aP/aY  & AwA  & LAD \tabularnewline
\hline
Ratio  & 86.96  & 76.00  & \textbf{43.48} \tabularnewline
\hline
\end{tabular}\caption{The ratios (\%) of shared classes between different datasets and ILSVRC
2012. Clearly, our LAD has the lowest overlap ratio.}
\label{tab:overlap}
\end{table}

\section{Data Split \label{sec:Data-Split}}

\label{sec:dsplit} The split of seen/unseen classes significantly
influences the performance of zero-shot learning methods. In previous
datasets such as CUB, SUN, aP/aY and AwA, only one split
of seen/unseen classes is specified for testing zero-shot algorithms.
However, due to the distinctive correlations of the classes, it is
not reliable nor convincing to evaluate the performance of algorithms
on the only one split. Hence, we propose a set of splits of seen/unseen
classes for zero-shot learning on our dataset. We adopt the idea of
five-fold cross validation to split the seen/unseen classes. Particularly,
we shuffle these classes and divide them into 5 folds. Each fold includes
20\% classes from every super-class. Each fold is used as the unseen
classes (20\%) in each split, and the rest are seen classes (80\%).
In this way, we obtain 5 random splits of seen/unseen classes to evaluate
the performance of zero-shot learning on our dataset.

We advocate others to evaluate their ZSL methods on each super-class
individually. It means that the data (images, labels, attributes)
of each super-class should be used separately. The performance on
each super-class should be the average value on all 5 splits. For
easy comparison, the average recognition accuracy on all super-classes can be used
as the general performance on our dataset. In experiments, we will
provide the evaluation of seven state-of-the-art ZSL methods using
these splits under the inductive setting.

For supervised learning, we randomly select 70\% data from each class
as training (train+validation) data and the rest 30\% are testing
data. These splits will be released along with our dataset.

\section{Methods and Experiments\label{sec:Experiments}}

This section will compare the state-of-the-art methods and conduct
the experiments under different settings on our dataset. In particular,
we consider the supervised learning (Sec. \ref{subsec:Supervised-Learning}),
zero-shot learning (Sec. \ref{subsec:Zero-shot-Recognition-by}),
zero-shot generation (Sec. \ref{subsec:Zero-shot-Generation}).

\subsection{Supervised Learning\label{subsec:Supervised-Learning}}

Though our LAD is designed as the testbed for zero-shot learning,
we can still validate the LAD in the standard supervised setting.
In particular, we provide several baseline results of the supervised
learning of objects and attributes. In each class, we have labeled
training data and unlabeled testing data (the split refers to Sec.
\ref{sec:dsplit}). We also show that our LAD is large enough and
has sufficient images to train the state-of-the-art deep architecture
\textendash{} ResNet.

\subsubsection{Object Recognition}

\textcolor{black}{
\noindent We use the state-of-the-art object recognition models, \emph{namely},
Inception-V3 \cite{inception_v3} and ResNet \cite{he2016deep} to
recognize objects. We train the two models under two settings,
\emph{namely}, with pre-training on  ILSVRC 2012 and without pre-training.
}

\textcolor{black}{
The recognition accuracies of LAD and AwA datasets are shown in Tab. \ref{Tab:objrec}.
In terms of deep models, ResNet works better than Inception-V3 in most settings.
Note that there are 230 classes in LAD and 50 classes in AwA.  The chance levels on the two datasets are 0.43\% and 2\% respectively.
However, the recognition accuracy on LAD is close to, even higher than, that on AwA. This phenomenon hints that more images and classes are beneficial to train deep models.
}

\textcolor{black}{
Generally speaking, the pre-training brings significant improvement of recognition accuracies for both two datasets and two models.
Averagely, on AwA dataset, the pre-training brings  38.79\% increase of recognition accuracy. However, the increase is only 26.79\% on our LAD dataset. This gap means that AwA dataset shares more classes with ILSVRC 2012 dataset.
}

\begin{table}
\centering{} %
\begin{tabular}{c|c|c|c|c}
\hline
 & \multicolumn{2}{c|}{w/o. Pre-training} & \multicolumn{2}{c}{w. Pre-training}\tabularnewline\hline
 & ResNet  & Inception-v3  & \multicolumn{1}{c|}{ResNet} & Inception-v3 \tabularnewline\hline
AwA  & 49.97 & 46.26 & \multicolumn{1}{c|}{86.51} & 87.29 \tabularnewline\hline
LAD  & 66.78  & 44.08  & \multicolumn{1}{c|}{84.92} & 79.52 \tabularnewline
\hline
\end{tabular}\caption{Object recognition accuracies (\%) on two datasets. w. means ``with'', and w/o. means ``without''.
Clearly, the pre-training on ILSVRC 2012 brings larger performance increase (averagely 38.79\%) for AwA than our
LAD (averagely 26.79\%). This result also denotes that AwA shares more classes with ILSVRC 2012.}
\label{Tab:objrec}
\end{table}

\vspace{0.04in}

\subsubsection{Attribute Recognition }

We also consider the task of recognizing different attributes. We
use the Inception-v3 features (without fine-tune) to learn attributes.
The images that belong to each class are randomly split into 70\%
training and 30\% testing data. The class-level attributes are binarized
to be 0 or 1, then the attribute recognition is a binary classification task. We train the Support
Vector Machine (SVM) with Multilayer Perception Kernel to learn each
attribute of each super-class. The classification accuracy of each
attribute is reported as the metric of the performance of attribute
recognition.

\begin{figure}
\centering{}\includegraphics[width=0.43\textwidth]{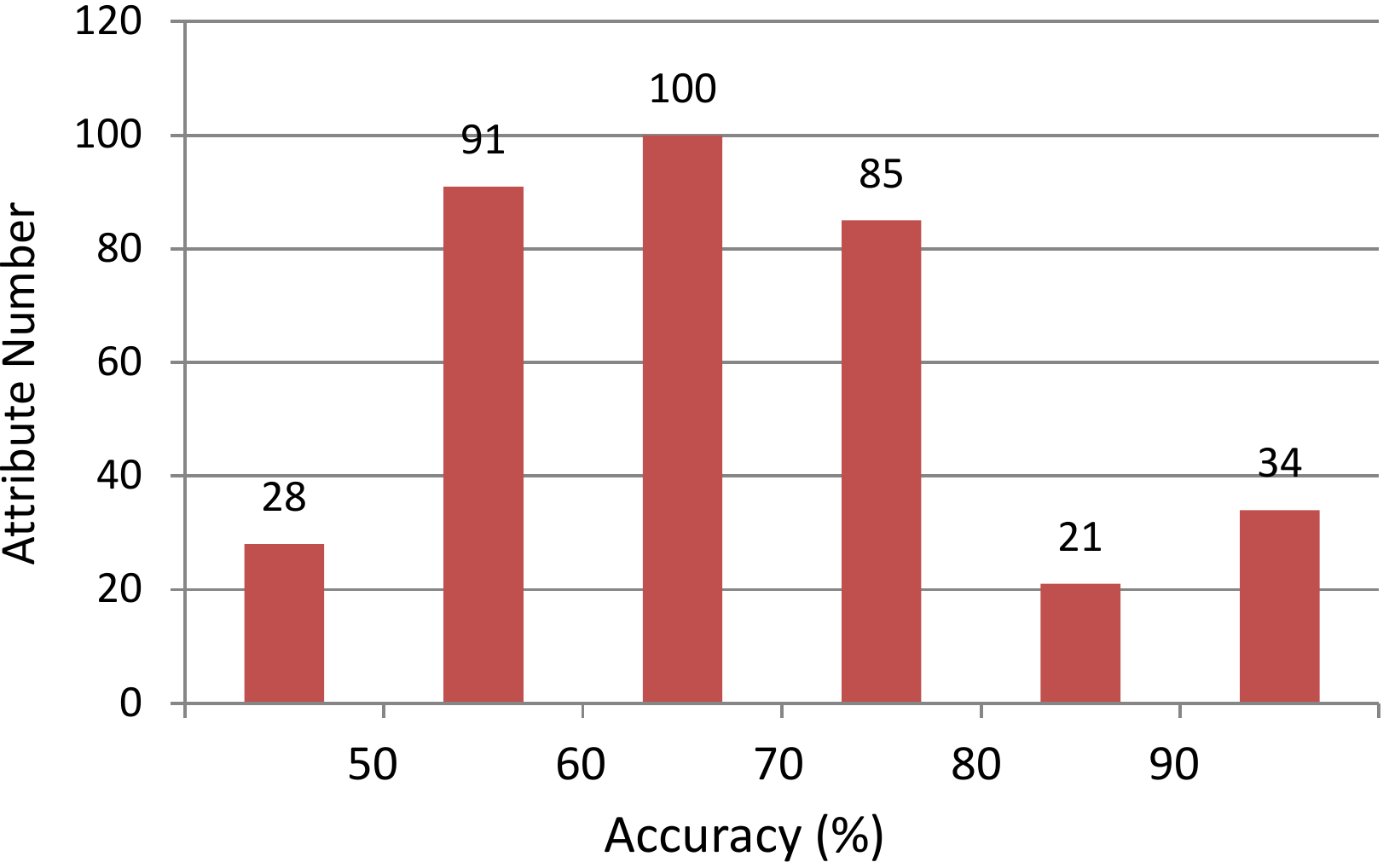} \caption{{}{}{The statistics of attribute recognition accuracies.}}
\label{fig:attrrecog}
\end{figure}

We histogram the attributes recognition accuracies into several intervals:
{[}0\%, 50\%), {[}50\%, 60\%), {[}60\%, 70\%), {[}70\%, 80\%), {[}80\%,
90\%), {[}90\%, 100\%{]}. As shown in Fig. \ref{fig:attrrecog}, the
recognition accuracies of most attributes of LAD are between 50\%
and 80\%. There are 28 attributes with lower than 50\% recognition
accuracy (even lower than the chance level), which means those attributes
are not well learned. We list some of those attributes in Tab. \ref{tab:attrlowacc}.
It is clear that most of those \textquotedbl{}hard\textquotedbl{}
attributes are about high-level semantics or those features that can
not be visually predicted. For example, \textquotedbl{}habit\textquotedbl{}
of animals, \textquotedbl{}safety\textquotedbl{} of vehicles, \textquotedbl{}aim\textquotedbl{}
of electronics and \textquotedbl{}feeling\textquotedbl{} of hairstyles
are about semantics which is hard to learn. Some attributes, e.g.
\textquotedbl{}appearance\_has soft skin\textquotedbl{} of animals,
\textquotedbl{}hardness\_is soft\textquotedbl{} of fruits, \textquotedbl{}material\_is
made of plastic\textquotedbl{} and \textquotedbl{}sound\_is quiet\textquotedbl{},
are also low-level perceptions, but other than vision. Thus the attributes
annotated in LAD is multi-modal. Currently, those attributes are difficult to
be predicted based on visual perception.

\begin{table*}
\centering{}%
\begin{tabular}{c|c|c}
\hline
Super-class  & hard-Attributes  & Explanation \tabularnewline
\hline
\hline
\multirow{2}{*}{Animal } & diet\_eats meat  & Whether the animal eats meat\tabularnewline
\cline{2-3}
 & habit\_is nocturnal  & whether the animal's habit is nocturnal\tabularnewline
\hline
\multirow{2}{*}{Fruits } & growth\_grow on trees  & whether the fruit grows on the tree\tabularnewline
\cline{2-3}
 & hardness\_is soft  & whether the hardness of the fruit is soft\tabularnewline
\hline
\multirow{2}{*}{Vehicles } & safety\_is safe  & whether the vehicle is safety\tabularnewline
\cline{2-3}
 & material\_is made of plastic  & whether the vehicle is made of plastic\tabularnewline
\hline
\multirow{2}{*}{Electronics } & aim\_is for display  & whether the electronics is designed for display\tabularnewline
\cline{2-3}
 & sound\_is quiet  & whether the sound of electronics is quite\tabularnewline
\hline
\multirow{2}{*}{Hairstyles } & feeling\_Is elegant  & whether the hairstyle gives the feeling of elegant\tabularnewline
\cline{2-3}
 & feeling\_Is sexy  & whether the hairstyle gives the feeling of sexy\tabularnewline
\hline
\end{tabular}{\tiny{}\caption{Attributes with low recognition accuracies. Please refer to the supplementary
material for the full list of ``hard'' attributes.}
\label{tab:attrlowacc}}
\end{table*}

\begin{table*}
\centering{}%
\begin{tabular}{c|c|c|c|c|c|c|c}
\hline
 & SOC  & ConSE  & ESZSL  & SJE  & SynC  & LatEm  & MDP \tabularnewline
\hline
Animals  & 50.76  & 36.87  & 50.15  & 61.89  & 61.60  & \textbf{63.92}  & 62.16 \tabularnewline
\hline
Fruits  & 40.01  & 29.77  & 37.23  & 46.39  & 51.42  & 44.23  & \textbf{56.40} \tabularnewline
\hline
Vehicles  & 56.98  & 37.48  & 45.75  & 63.00  & 54.89  & 60.94  & \textbf{65.09} \tabularnewline
\hline
Electronics  & 33.73  & 28.27  & 32.83  & 39.51  & 42.97  & 40.71  & \textbf{45.11} \tabularnewline
\hline
Hairstyles  & \textbf{42.45}  & 24.55  & 31.84  & 38.50  & 29.10  & 38.53  & 42.12 \tabularnewline
\hline
Average  & 44.79  & 31.39  & 39.56  & 49.86  & 48.00  & 49.67  & \textbf{54.18} \tabularnewline
\hline
\end{tabular}\caption{\label{tab:zsr} The performance (\%) of seven state-of-the-art ZSR
methods on our dataset.}
\end{table*}

\subsection{Zero-shot Recognition by Attributes\label{subsec:Zero-shot-Recognition-by}}

We propose LAD as the new testbed for zero-shot recognition. In particular,
as the sanity check, severn state-of-the-art zero-shot learning algorithms
are re-implemented; and their results are reported and compared in
this section. We use the data split proposed in Sec. \ref{sec:dsplit}.
For all methods, we use the ResNet feature extractor which is trained
on training images of ILSVRC 2012. We follow the inductive learning
setting, \emph{i.e.}, data from unseen classes are not available for
training.

\vspace{0.1in}

\noindent \textbf{Methods}. We compare seven state-of-the-art zero-shot
learning methods, including SOC \cite{palatucci2009zero}, ConSE \cite{norouzi2013zero},
ESZSL \cite{romera2015embarrassingly}, SJE \cite{akata2015evaluation},
SynC \cite{changpinyo2016synthesized}, LatEm \cite{xian2016latent},
MDP \cite{zhao2017zero}. SOC \cite{palatucci2009zero} learns the
mapping from image features to semantic output codes (semantic embeddings)
using seen classes. Then the learned mapping is used for predicting
the semantic output codes of images from unseen classes. ConSE \cite{norouzi2013zero}
maps the images into the semantic embedding space by the convex combination
of the class label embedding vectors. The benefit is that this method
does not need an extra training for unseen classes. ESZSL \cite{romera2015embarrassingly}
learns the bi-linear mapping function which maps both the image features
and semantic embeddings to the new space. SJE \cite{akata2015evaluation}
proposes to learn the compatibility function which measures the compatibility
between the the image features and semantic embeddings. The function
is trained on seen classes and tested on unseen classes. SynC \cite{changpinyo2016synthesized}
learns to synthesize classifiers for unseen classes by the linear combination
of classifiers for seen classes. LatEm \cite{xian2016latent} proposes
a new compatibility function which is a collection of bilinear maps.
These bilinear maps can discover latent variables. MDP \cite{zhao2017zero}
aims to learn the local structure in the semantic embedding space,
then transfer it to the image feature space. In the image feature
space, the distribution of unseen classes are estimated based on the
transferred structural knowledge and the distribution of seen classes.

We can roughly split these methods into two categories in term of
how the knowledge is transferred. The first one is mapping-transfer
(including SOC \cite{palatucci2009zero}, ConSE \cite{norouzi2013zero},
ESZSL \cite{romera2015embarrassingly}, SJE \cite{akata2015evaluation}
and LatEm \cite{xian2016latent}), which learns the mapping between
the image features and semantic embeddings on seen classes. Then,
the learned mapping is transferred to unseen classes for predicting
the labels. The second category (including SynC \cite{changpinyo2016synthesized}
and MDP \cite{zhao2017zero}) is about structure-transfer, which learns
the structural knowledge (relationship) between seen and unseen classes
in the semantic embedding space. Then the learned structural knowledge
is transferred to the image feature space for synthesizing classifiers
or estimating the data distribution of unseen classes.

Note that there are many zero-shot algorithms such as SS-Voc \cite{fu2016semi}
that heavily rely on the word vectors (e.g. Word2Vec\cite{mikolov2013distributed},
or GloVec\cite{GloVec}) of the class names. However, in LAD, the
class name is less informative to represent the whole data distribution
in the semantic layer, \emph{e.g.}, the ``fauxhawk'' class in the hairstyle
super-class. Therefore, for a more fair comparison, algorithms
that are heavily relying on word vectors have not be compared
here. The zero-shot recognition is conducted on each super-class separately.

Tab. \ref{tab:zsr} shows the zero-shot recognition accuracies of
different methods. In general, MDP achieves the best performance (54.18\%
averagely). This result is higher than the runner-up (SJE) by 4.32\%.
Among all the super-classes, most algorithms can easily achieve relative
high performance on the super-classes of ``Animals'' and ``Vehicles''.
This is reasonable, since these two super-classes include very common
objects/concepts which have been widely collected in the ImageNet
dataset. Hence, the feature extractor pre-trained on ILSVRC 2012 may
work better on the two super-classes than the other ones. Almost all
the algorithms have relative low performance on ``Hairstyles'' super-class,
even although we only split 6 unseen classes. More impressively, the
SOC \cite{palatucci2009zero} proposed in 2009 can beat all the other
methods on ``Hairstyles'' super-class.

\subsection{Zero-shot Generation\label{subsec:Zero-shot-Generation}}

We also conduct experiments for the zero-shot generation task. This
task aims to generate images of unseen classes, \emph{i.e.}, those
with novel attribute representations. Note that this task is extremely
challenging due to both the diverse and fine-grained classes in LAD
and the high-semantic multi-modal. In particular, we conduct the experiments
on the ``Animals'' super-class.

\vspace{0.1in}

\begin{figure}
\centering \includegraphics[width=0.48\textwidth]{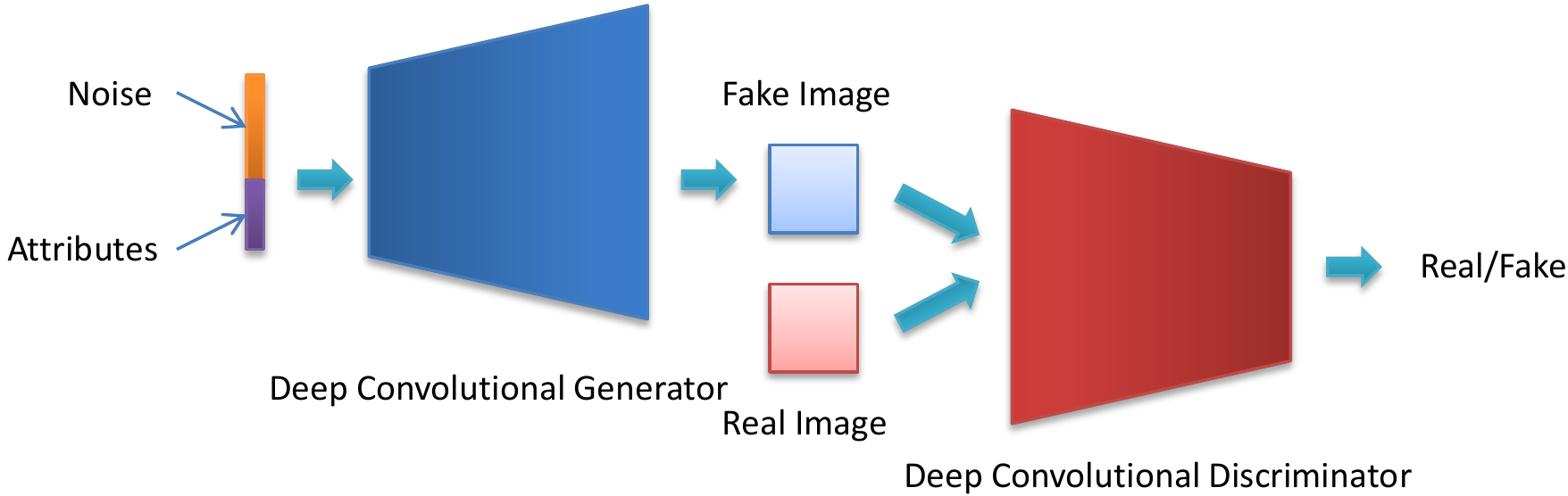}
 \caption{{}{The model structure for zero-shot generation.}}
\label{fig:dcganattr}
\end{figure}

\noindent \textbf{Methods}.  Based on the Deep Convolutional Generative
Adversarial Networks (DCGAN) \cite{radford2015unsupervised}, we introduce
the condition by concatenating the condition vector and the noise
vector. In DCGAN, the generator and discriminator are two deep convolutional networks which
have 4 convolution layers. Refer to \cite{radford2015unsupervised} for the detailed convolutional structure.  Both input and output images are reshaped
to $64\times64\times3$. The attributes of each image serves
as the condition. In this way, the learned DCGAN can generate images
conditioned on attributes. \textcolor{black}{We illustrate the model structure  in Fig. \ref{fig:dcganattr}.  }

\vspace{0.1in}

\begin{figure}
\centering \includegraphics[width=0.47\textwidth]{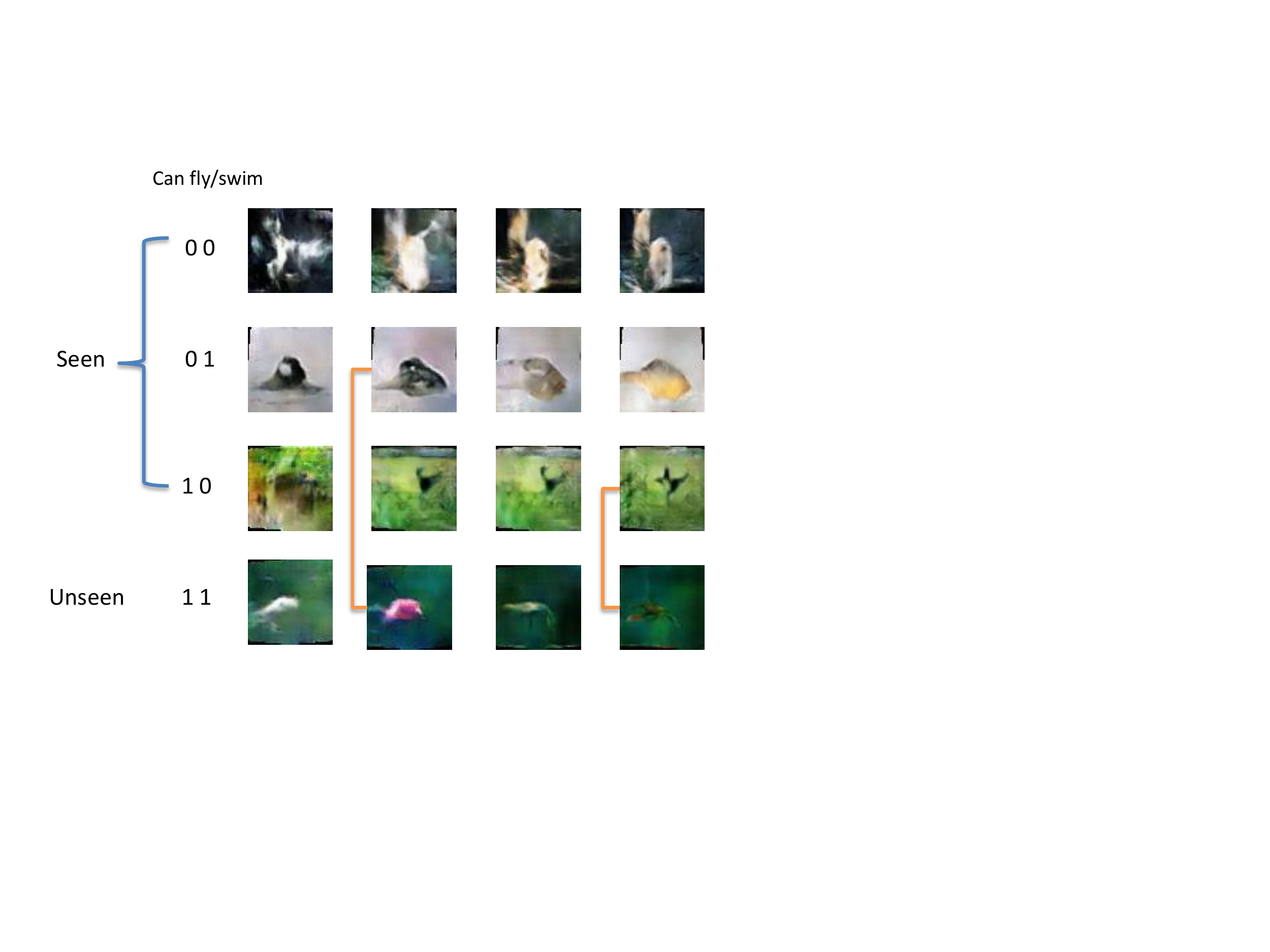} \caption{{\small{}{}{}{}{}{}Results of zero-shot generation. The seen
classes have the attributes 00, 01 and 10, which denote whether it can
fly or swim. The unseen class has the attribute representation of 11. From left to right, we display the generated images with more training iterations. }}
\label{fig:zsg}
\end{figure}

\noindent \textbf{Results}. The animals in our dataset are divided
into 3 folds in term of their attributes \textendash{} ``can swim''(1/0)
and ``can fly'' (1/0), namely, 00, 01, 10. Then we train a GAN conditioned
on the two attributes.
\textcolor{black}{ The trained model is used to generate new images
with both seen attribute representations and the unseen one (with the attribute representation ``11''). As shown in Fig. \ref{fig:zsg},
objects with the particular seen attributes can be generated. From
the left to right, more clear images are generated with more training
iteration.
The objects in the first three rows look like ``monkey'', ``fish'' and ``bird'' respectively.
In the 4th row, the generated unseen object
looks like a ``fish'' in the 2nd stage (column). Later, the ``wings'' are observed in the 3rd stage.
This result means that the novel object
can be generated based on novel the attribute representation. Note that the distribution
of the training seen images in LAD is very diverse; and thus it is intrinsically
a very challenging task to generate the images of novel unseen classes.
}

\vspace{0.1in}

\begin{figure}
\centering \includegraphics[width=0.44\textwidth]{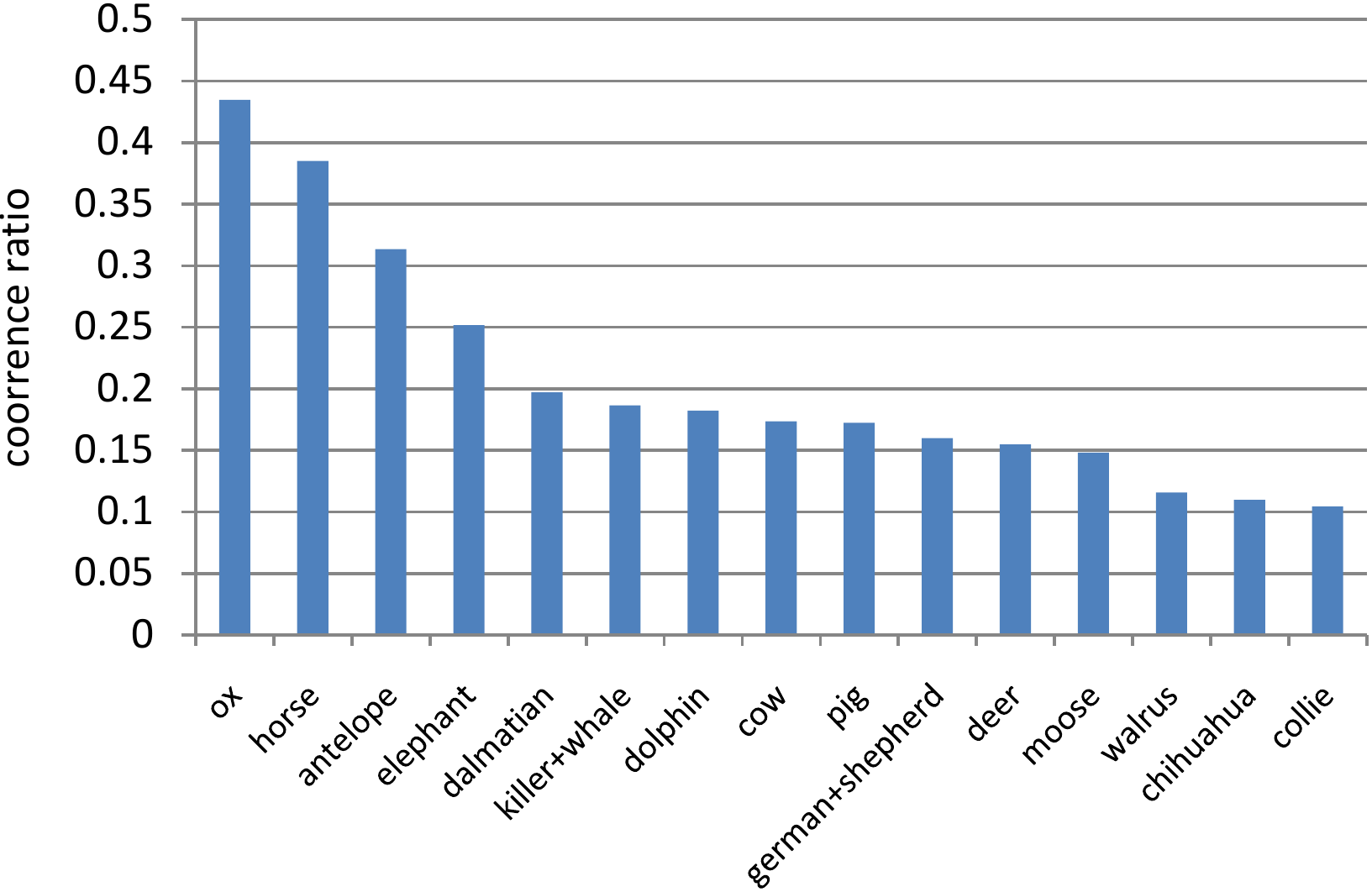} \caption{{}{The co-occurrence bias on AwA dataset. This figure shows the top 15 animal classes with high co-occurrence ratios of \textquotedbl{}persons\textquotedbl{}.}}
\label{fig:distbias}
\end{figure}

\begin{figure}
\centering{}\centering \includegraphics[width=0.47\textwidth]{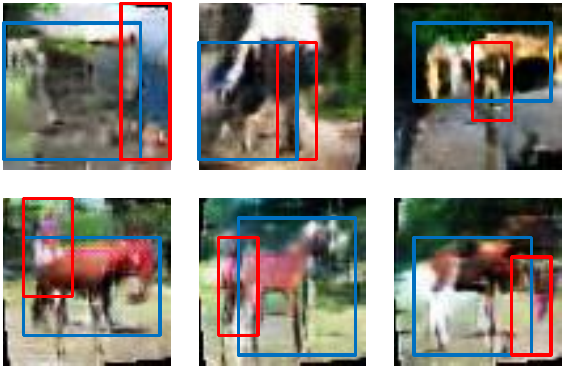}
\caption{Visualization of the co-occurrence bias in AwA. The upper three images are synthesized images of ``ox'', and the below images are those of ``horse''. Blue boxes are animals,
while red boxes are ``persons'', i.e., bias.}
\label{fig:bias}
\end{figure}

\noindent \textbf{Analysis of Co-occurrence Bias in AwA.} We also
study the co-occurrence bias in previous datasets. We analyze and
visualize the influence of the co-occurrence bias on the learning of a particular concept. Specifically, we
first count the co-occurrence of \textquotedbl{}person\textquotedbl{}
and different animals in AwA dataset. We use YOLO \cite{redmon2016yolo9000},
which is pre-trained on MS-COCO \cite{lin2014microsoft}, to detect
\textquotedbl{}person\textquotedbl{} in each image. We only count
those appeared \textquotedbl{}person\textquotedbl{} with high ($>70\%$)
probability. As shown in Fig. \ref{fig:distbias}, 15 classes in AwA
have large ($>10\%$) co-occurrence ratio of \textquotedbl{}person\textquotedbl{}.
Clearly, the co-occurrence bias of \textquotedbl{}person\textquotedbl{}
is serious in AwA. To visualize the influence of the co-occurrence
bias, we use the GAN, which can well capture the data distribution,
to learn the concepts ``ox'' and ``horse''. Fig. \ref{fig:bias}
illustrates the synthesized images of the two animals. Except animals
in blue boxes, the ``persons'' in red boxes are also synthesized
in the image. \textcolor{black}{These synthesis results illustrate that the co-occurrence bias
may cause the mis-learning of a particular concept. Thus during the construction of LAD, we reduce the
correlation bias by filtering multi-object images, \emph{i.e.}, we
preserve images with only one foreground object.}

\section{Conclusion\label{sec:Conclusion}}

In this paper, we present a Large-scale Attribute Dataset (LAD) for zero-shot
learning. Many attributes about visual, semantic and subjective properties
are annotated. We re-implement and compare seven state-of-the-art
methods and report the performance as the baselines on our dataset.
Experiments show that our dataset is still a challenging for zero-shot
learning. \textcolor{black}{This proposed LAD has been used as the the benchmark dataset for Zero-shot Learning Competition in AI Challenger.} In the future, more investigation should be devoted to semantic and subjective attributes for deeper understanding of images.

\small{
\bibliographystyle{ieee}
\bibliography{egbib}
}

\end{document}